\documentclass[10pt,twocolumn,letterpaper]{article}

\usepackage[pagenumbers]{cvpr} % To force page numbers, e.g. for an arXiv version

\usepackage[dvipsnames]{xcolor}

\usepackage{csquotes}
\usepackage{enumitem}
\usepackage{csquotes}
\usepackage{soul}
\usepackage{bm}

\definecolor{cvprblue}{rgb}{0.21,0.49,0.74}
\usepackage[pagebackref,breaklinks,colorlinks,citecolor=cvprblue]{hyperref}

\def\paperID{8916} % *** Enter the Paper ID here
\def\confName{CVPR}
\def\confYear{2024}

\title{Novel View Synthesis with View-Dependent Effects from a Single Image}

\author{%
  Juan Luis Gonzalez Bello \\
  {\tt\small juanluisgb@kaist.ac.kr}
  \and
  Munchurl Kim \\
  {\tt\small mkimee@kaist.ac.kr}
  \and
  Korea Advanced Institute of Science and Technology (KAIST) \\
}

\begin{document}
% \maketitle

\twocolumn[{
\renewcommand\twocolumn[1][]{#1}
\maketitle
\begin{center}\centering
  \includegraphics[width=0.99\textwidth]{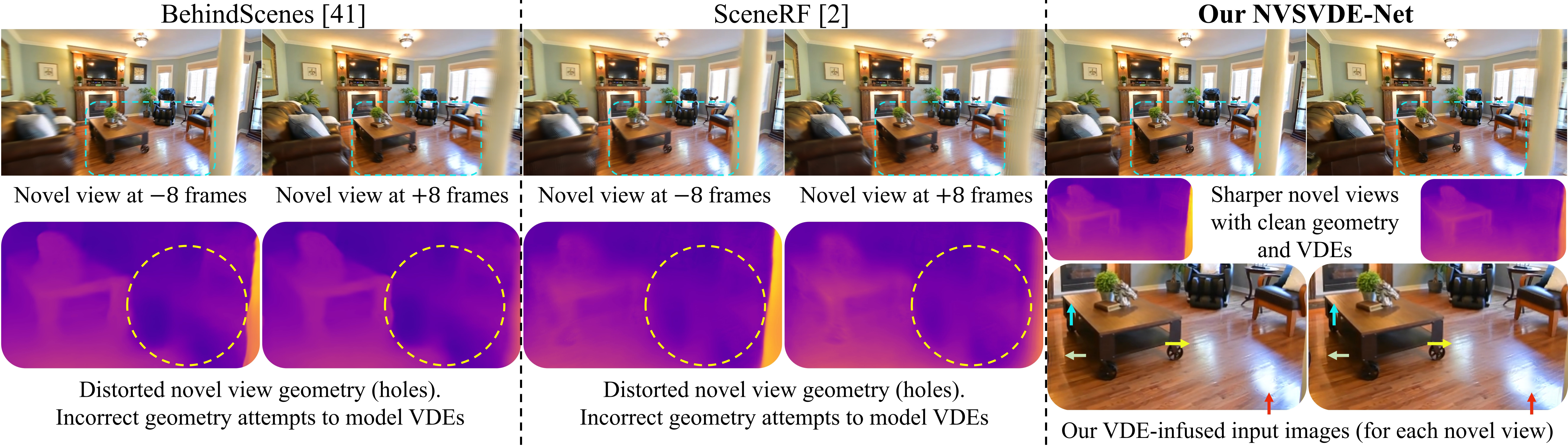}
  \vspace*{-2.5mm}
  \captionof{figure}{Comparison of single-view NVS. Visit our project site \url{https://kaist-viclab.github.io/monovde-site/}.
  }
  \label{fig:opening_img}
\end{center}
}]

\begin{abstract}

In this paper, we \textit{firstly} consider view-dependent effects into single image-based novel view synthesis (NVS) problems. For this, we propose to exploit the camera motion priors in NVS to model view-dependent appearance or effects (VDE) as the negative disparity in the scene. By recognizing specularities \enquote{follow} the camera motion, we infuse VDEs into the input images by aggregating input pixel colors along the negative depth region of the epipolar lines. Also, we propose a `relaxed volumetric rendering' approximation that allows computing the densities in a single pass, improving efficiency for NVS from single images. Our method can learn single-image NVS from image sequences only, which is a completely self-supervised learning method, for the first time requiring neither depth nor camera pose annotations.
We present extensive experiment results and show that our proposed method can learn NVS with VDEs, outperforming the SOTA single-view NVS methods on the RealEstate10k and MannequinChallenge datasets.
\end{abstract}    

% \begin{figure}[t]
%   \centering 
%   \includegraphics[width=0.48\textwidth]{Figures/splash.jpg}
%   \vspace*{-6mm}
%   \caption{Comparison of novel view synthesis with view-dependent effects. See Supplemental and \url{https://shorturl.at/ltJT7} for realistic and animated renderings.}
%   \vspace*{-4mm}
% \end{figure}

\enlargethispage{\baselineskip}
\section{Introduction}
\label{sec:intro}

Novel view synthesis (NVS) is a fundamental computer vision task that aims to generate new views of the input scene from arbitrarily different camera positions. Recent advances in NVS have shown that view-dependent effects (VDE), which increase the realism and perceived quality of the novel views, can also be learned and incorporated into the rendering pipelines. Particularly, NeRF \cite{nerf} has demonstrated that radiance fields can be effectively learned by a multi-layer perceptron (MLP) that allows rendering geometry and VDEs from multi-view captures. 

% \footnote{\textdagger Corresponding author.}

NeRF relies purely on multi-view consistency and cannot exploit prior knowledge, such as textures and depth cues common across natural scenes. This limits NeRF when a few or only one single view is available. On the other hand, to leverage the 3D prior knowledge in multi-view datasets, PixelNeRF \cite{pixelnerf} proposed to train an MLP that takes as inputs the spatial locations, the viewing directions, and the pixel-aligned deep features to generate the colors and densities in radiance fields. However, PixelNeRF is still incapable of learning the ill-posed task of directly mapping input image pixels to VDEs. Other works, such as single-image MPIs \cite{single_view_mpi}, MINE \cite{mine}, BehindTheScenes \cite{behindScenes}, and SceneNeRF \cite{scenerf} have also proposed single-view-based NVS, but cannot model VDEs. 

VDEs depend on the material's reflectance, which is a function of the material properties and the light's angle of incidence. Learning such material properties and light sources from a single image is a very ill-posed problem. While MLP \cite{nerf} and spherical-harmonic-based \cite{plenoctrees, plenoxels, nex} techniques to encode VDEs are effective when learning from multiple input images, they are still constrained when learning from single images. Instead,\textit{ for the first time}, to tackle the estimation of VDEs for single-view-based NVS, we propose to rely on the contents of the single images and estimated (during training) or user-defined (during test time) camera motions to estimate photo-metrically realistic view-dependent effects. Our main contributions are:
\begin{itemize}
\vspace*{0.5mm}
\item We \textit{firstly} propose a single view NVS method with VDEs, called NVSVDE-Net. By recognizing camera motion priors and negative disparities govern view-dependent appearance, we model VDEs as the negative disparities in the scene induced by the target camera motion.

\vspace*{0.5mm}
\item The NVSVDE-Net is the \textit{first} to be trained in a completely self-supervised manner in the sense that neither depths nor pose annotations are required. While other methods rely on given depths and/or camera poses, the NVSVDE-Net learns \textit{only} from image sequences. During test, only one single image input is required for the NVSVDE-Net to render novel views with VDEs.

\vspace*{0.5mm}
\item A novel `relaxed volumetric rendering' approximation method is proposed, allowing fast and efficient rendering of novel views from a single image. The NVSVDE-Net's rendering pipeline well approximates volumetric rendering by a single pass of a convolutional (or transformer-based) backbone, probability volume adjustment MLP blocks, and a sampler MLP block.
\end{itemize}

Additionally, to better learn from image sequences only, we introduce a new coarse-to-fine camera pose estimation network as a secondary contribution.
\enlargethispage{\baselineskip}
\section{Related Works}
\label{sec:related_works}

In contrast with classical techniques for single view NVS, which require user input or rely on hard-coded domain-specific assumptions \cite{classic_tourin, classic_photopop}, deep-learning-based approaches can leverage image representations that are common across scenes to render novel views automatically.  
We distinguish between two kinds of deep-learning-based NVS methods from single images: one kind that requires an additional pre-computed depth map \cite{single3dphoto, 3dkenburns, adampi} and the other kind that only uses a single image during test time \cite{deep3d, single_view_mpi, pixelnerf, mine, behindScenes, pose_guided_diffusion, scenerf, vq3d}. Our proposed method lies in the second category, with the remarkable exception of not requiring ground truth (GT) depths and GT camera poses during training. Moreover, we focus on static forward-facing scenes, not object-centric 360-degree rendering \cite{vision_nerf} or modeling motion of dynamic objects \cite{dynibar}.

\subsection{MPI-Based Single View NVS}

The multi-plane image representation (MPI) \cite{mpi} maps single or multiple images into a camera-centric layered 3D representation. Each input image pixel $p=(u,v)$ is mapped to $D$-number of colors and opacities.
New views are rendered by sampling the MPI colors and opacities from novel camera poses and by aggregating them via volumetric rendering \cite{volume_render}. 
% Planar affine transformations are used to project the MPI's planes into the target camera pose, which aligns the points along the target rays for alpha compositing. 
In \cite{single_view_mpi}, Tucker and Snavely proposed single-view NVS with MPIs. Estimating densities and colors in MPIs is challenging and does not allow for modeling view-dependent effects, as the MPI's colors are not a function of the viewing directions. Furthermore, Sampling distances in MPIs depend on intersections with multi-image planes, potentially hindering high-quality rendering. On the other hand, MINE \cite{mine} borrows from \cite{nerf} and \cite{mpi} by adopting an MPI that is trained and queried via NeRF-like strategies, where the MPI representation is decoded one depth plane at a time. However, MINE still cannot model VDEs, and its sequential decoding increases computational complexity.

% TODO check if mine needs supervision
In contrast, our method explicitly models VDEs and approximates volumetric rendering by relaxing alpha compositing and obtaining refined sample distances at the target camera view for fine-grained rendering. Furthermore, we do not rely on ORB-SLAM2 predicted camera poses and sparse 3D point clouds for supervision.

\subsection{NeRF-based Single View NVS}

In Neural Radiance Fields (NeRF) \cite{nerf}, Mildenhall \etal introduced a method that effectively maps 3D coordinates and viewing directions to color and density values using an MLP for volumetric rendering. 
% This technique has proven effective for modeling 3D geometry and view-dependent appearance from carefully captured scenes.
% NeRF's success is also attributed to its coarse-to-fine rendering strategy and positional encoding of 3D coordinates. 
However, NeRF has limitations: Besides overfitting to a single scene, it needs several reference views with accurate camera poses and long processing times. Nevertheless, recent advances have dramatically reduced neural rendering train and test times \cite{mobilenerf, 3Dgaussians}.

PixelNeRF \cite{pixelnerf} is a cross-scene generalizable variant of NeRF, sacrificing rendering quality for scene generalization. It maps not only 3D ray points but also projected deep features into color and opacity values in radiance fields. Even when viewing directions are incorporated into PixelNeRF's MLP, the lack of a view-dependent inductive bias prevents PixeNeRF from generating VDEs.

Wimbauer \etal extended PixelNeRF in BehindScenes \cite{behindScenes} by sampling colors from epipolar lines and processing densities with an MLP head for each point in the target ray. BehindScenes is designed to focus on 3D geometry but still lacks a mechanism for modeling VDEs as colors are projected between the minimum and maximum distance bounds in the target view. SceneRF \cite{scenerf} is also built on PixelNeRF but incorporates a probabilistic ray sampling strategy and a Spherical U-Net, along with depth penalties \cite{monodepth, monodepth2}. Their sampling strategy uses Gaussian mixtures and requires multiple forward passes for rendering. Like NeRF\cite{nerf}, the MLPs in PixelNeRF \cite{pixelnerf}, BehindScenes \cite{behindScenes}, and SceneRF \cite{scenerf} independently process each sample, amounting to a considerable computational complexity. In contrast, our approach with fine-grained ray sampling and relaxed volumetric rendering allows for all opacity estimations in a \textit{single pass}, yielding efficiency gains.

\enlargethispage{\baselineskip}
\subsection{Generative Single-View NVS}

NVS from a single image has also benefited from new advances in generative models. In \cite{ren2022look}, Ren \etal used an autoregressive transformer with a VQ-GAN \cite{esser2021taming} decoder to model long-term future frames. Similarly, Rombach \etal proposed GeoGPT \cite{geogpt} where they showed that autoregressive transformers can implicitly learn 3D relations between source and target images. In \cite{pose_guided_diffusion}, Tseng \etal proposed a diffusion model \cite{ddpm} with epipolar attention blocks at its bottleneck that learn to infuse source view information along the epipolar lines into the target synthetic view during the diffusion process. In VQ3D \cite{vq3d}, Sargent \etal proposed to learn 3D-aware representations and generation from ImageNet \cite{imagenet} using transformer-based \cite{vit} autoencoders to map images and latent vectors into a tri-plane representation for neural rendering \cite{chan2022efficient}. Pre-trained DPT \cite{dpt} depths and various adversarial losses are used to train VQ3D. VQ3D is limited to render low-resolution views (256$\times$256) with constrained viewpoints more related to object-centric rendering.

Even when generative models aid NVS from a single image in generating novel views with very large baselines, they suffer from severe inconsistencies due to the stochastic nature of generative models. In addition, they suffer from large computational requirements and inference times that still make them impractical for single-view NVS.

\enlargethispage{\baselineskip}
\subsection{View Dependent Effects}
View-dependent effects (VDE) in NVS from multiple images have been achieved by incorporating viewing directions into the scene representations. In the case of NeRFs \cite{nerf, mipnerf}, viewing directions are fed into the late stages of the MLPs, while in other works \cite{plenoxels, plenoxels, nex} view-dependent effects are modeled by spherical harmonics which map viewing directions to intensity changes.

Even though these methods show SOTA view-dependent effects, they have the severe limitation of not generalizing cross-scenes and requiring several reference frames for inference. Moreover, fitting new scenes is a time-consuming optimization process that can take several hours. 

In contrast, we introduce a new rendering pipeline, the NVSVDE-Net, which learns to approximate volumetric rendering for estimating novel views and view-dependent effects (VDE) from videos without requiring camera pose or depth labels. Trained in a self-supervised manner from image sequences, our NVSVDE-Net can render novel views with VDEs during inference, even from unseen single-image inputs. This marks the first instance of showcasing VDEs estimated from single images, leveraging local context and camera motion priors.

% In contrast, we propose a novel rendering pipeline that approximates volumetric rendering to estimate geometries and view-dependent effects (VDE) from videos without camera pose or depth labels. It is worth noting that our NVSVDE-Net is trained in a self-supervised manner from image sequences, but during inference it can render novel views with VDEs from an unseen single-image input. That is, our NVSVDE-Net is a single-image-based NVS model. For the first time, we show VDEs estimated from single images by exploiting local context and camera motion priors.
\enlargethispage{\baselineskip}
\section{Proposed Method}
\label{sec:method}

Volumetric rendering \cite{volume_render} synthesizes novel camera views by traversing rays $\bm{r}$ that originate in the target view camera center into a 3D volume of colors $\bm{c}$ and densities $\sigma$. The continuous volumetric rendering equation (VRE) is
\begin{equation}\label{eq:vre}
    C\bm{(r)} = \int^{t_f}_{t_n} \mathcal{T} (t)\sigma(t) \bm{c}(t) dt, 
\end{equation}
where the accumulated transmittance $\mathcal{T}(t)$ indicates the probability of $\bm{r}$ traveling between the near distance bound $t_n$ and ray distance $t$ without hitting a particle. On the other hand, the density $\sigma(t)$ is understood as the probability of $\bm{r}$ hitting a particle exactly at $t$. $t_f$ is the far distance bound.

In NeRF \cite{nerf}, a large number of samples along the ray is considered to discretely approximate the VRE by Monte-Carlo sampling while the processing blocks, such as an MLP or convolutional layers, compute $\sigma_i$ and $\bm{c}_i$ for each $i^{th}$ sample along the ray. For this reason, the methods inspired in NeRF \cite{pixelnerf, mine, behindScenes, scenerf} tend to yield very slow rendering times as each sample requires an independent forward pass.

\begin{figure*}
  \centering 
  \includegraphics[width=0.99\textwidth]{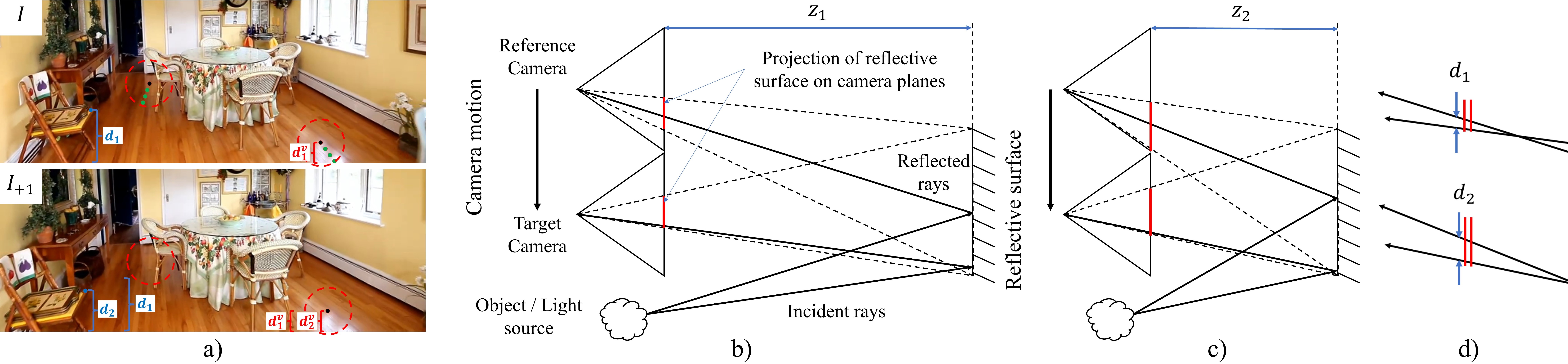}
  \vspace*{-3mm}
  \caption{
  VDEs `follow' camera motion (no disparities). (a) We synthesize target VDEs (black dots) by re-sampling $I$ along the negative depth epipolar line region (green dots). 3D geometry does present disparities between views (blue dots). (b) VDEs present disparities relative to their reflective surfaces in the opposite direction than the projection of the reflective surface itself. (c) and (d) VDE disparity due to novel camera poses is proportional to the reflective surface disparity. The closer the reflective surface, the larger the VDE disparity.}
  \label{fig:vde_follow}
\end{figure*}

\subsection{Relaxed Volumetric Rendering}

Estimating the $\bm{c}_i$ and $\sigma_i$ for volumetric rendering is not a trivial task. The complexity of volumetric rendering increases when a few or a single observation is available, as it becomes a one(pixel color)-to-many(densities and colors) mapping. For this reason, recent works such as \cite{behindScenes} ease the network burden by only predicting density values (by several passes of an MLP head) and using colors directly projected from the input image $I$. 

To better incorporate a 3D inductive bias into our models, we propose to relax further the VRE in Eq. (\ref{eq:vre}) by directly modeling the `ray point weights' also known as $\mathcal{T} (t)\sigma(t)$. Eq. (\ref{eq:vre}) can further be relaxed by using the projected colors of a VDE-infused input image $I^v_c$ from camera-view $c$ instead of using the estimated colors $\bm{c}_i$. 

For each point in the ray, VDE-infused colors $I^v_c$ and depth probabilities or `weights' $D^P$ are epipolar-projected into target camera $c$, and they are integrated to generate the final color estimate $I'_c(\bm{p})$. Note that there is a ray $\bm{r}$ for each pixel location $\bm{p}=(u,v)$. A novel view $I'_c$ by our relaxed volumetric rendering is given by
\begin{equation} \label{eq:vre_approx}
I'_c(\bm{p}) = \sum^{N-1}_{i=0} D^P_i(\bm{p}) I^v_c(g(\bm{p}, t_i, R_c|\bm{t}_c, K)),
\end{equation}
where $R_c$ and $\bm{t}_c$ are the target camera rotations and translations, $K$ is the camera intrinsics, and $N$ is the number of samples in the ray $\bm{r}$. Following the literature \cite{dorn, falnet}, $t_i = t_n(t_f/t_n)^{1-i/(N-1)}$ exponentially distributes the sample points in the target rays such that there are more sampling points for the closer depths and fewer for the far-away distances. $g(\cdot)$ is the epipolar projection function that outputs the pixel coordinate $g(\cdot)=(u',v')$, allowing to sample colors from $I^v_c$ at $(u',v')$ by bilinear interpolation. See \textit{Supplemental} for more details on $g(\cdot)$. Finally, $D^P_i(\bm{p})$ is the $i^{th}$ channel of the projected depth probability volume, which approximates $ \mathcal{T} (t)\sigma(t)$ in Eq. (\ref{eq:vre}) for our relaxed volumetric rendering, and is given by
\begin{equation} \label{eq:vre_Dp}
D^{P}(\bm{p}) = \sigma \left(\{D^L_i(g(\bm{p}, t_i, R_c|\bm{t}_c, K))\}^{N-1}_{i=0} \right),
\end{equation}
where $\sigma(\cdot)$ denotes the channel-wise softmax operator and $\{D^L_i(\cdot)\}^{N-1}_{i=0}$ denotes the channel-wise concatenation of $D^L_i(\cdot)$. $D^{L}$ is the estimated depth logit volume, which describes the scene's geometry seen from the input or source camera view. Additionally, a depth estimate $\hat{D}$ can also be drawn from $D^{L}$ by a dot-product with all $t_i$ by
\begin{equation} \label{eq:depth_hat}
\hat{D}(\bm{p}) = \{t_i\}^{N-1}_{i=0} \cdot \sigma(D^L(\bm{p})).
\end{equation}

Contrary to MPIs \cite{mpi, single_view_mpi}, our approximated volumetric rendering allows us to efficiently and uniformly sample the target view rays, instead of relying on the intersections of the target rays into the multi-image planes. 

\begin{figure*}[t]
  \centering 
  \includegraphics[width=1.0\textwidth]{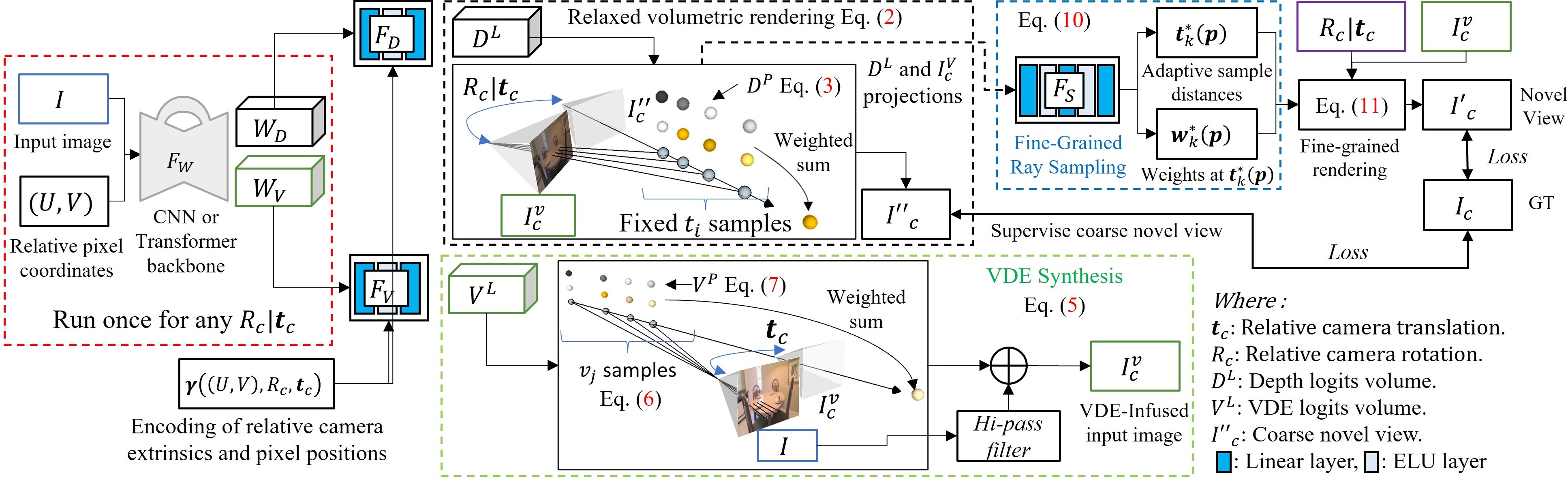}
  \vspace*{-7mm}
  \caption{NVSVDE-Net architecture. NVSVDE-Net models VDEs as negative scene disparities under the target camera motion $R_c|\bm{t}_c$. Novel views are estimated in two stages. Firstly with coarse fixed ray samples $t_i$, then with refined adaptive sampling distances $t^*_k(\bm{p})$.}
  \label{fig:overall_arch}
\end{figure*}

\enlargethispage{\baselineskip}
\subsection{Synthesis of View-Dependent Effects}
\label{sec:method_vde}

View-dependent effects (VDE), such as the glossy reflections depicted in the red circles of Fig. \ref{fig:vde_follow}-(a), have almost `no disparities.' That is, they seem to stay in similar image regions in both the current $I$ and the next frame $I_{+1}$. In other words, VDEs `follow' the camera motions relative to their reflective surfaces as conceptually depicted in Fig. \ref{fig:vde_follow}-(b). While the reflective surface projection into the reference camera (red line) `moves' upwards in the target camera, the reflected ray (or \textit{view-dependent effect}) moves downwards with respect to the reflective surface.

We propose to exploit this strong, yet simple and effective prior in VDEs by generating a target view-dependent appearance (black dots in Fig. \ref{fig:vde_follow}-(a) by re-sampling the pixels that `follow' the camera motion (green dots in Fig. \ref{fig:vde_follow}-(a)). This operation is equivalent to a weighted sum of pixels along the negative depth region of the epipolar line. Optionally, the same effect can be achieved by keeping a positive VDE depth and inverting the relative camera motion between the source and target views. We then define the VDE-infused input image $I^v_c$ at the viewing direction of camera $c$ as
\begin{equation} \label{eq:vde_syn}
I^{v}_{c}(\bm{p}) = I_{H} + \sum_{j=0}^{N_v-1} V^P_j(\bm{p}) I\left(g(\bm{p}, 1/v_j,  \mathbf I|\bm{t}_c, K)\right),
\end{equation}
where $I_{H} = I - I * k_{5\times5}$ roughly contains the high-frequency details of I by subtracting the low-pass box-kernel-filtered $I * k_{5\times5}$ from $I$. $I_{H}$ aids in generating VDEs while preserving structural details. $N_v$ is the number of VDE samples from $I$. 
% Like $D^P$ in Eq. \ref{eq:vre_approx}, 
$V^P_i$ are the projected VDE probabilities or `weights' at the location $\bm{p}$. Note that the identity matrix $\mathbf I$ (no 3D rotation) in Eq. (\ref{eq:vde_syn}) is used instead of $R_c$, as pure 3D rotations cannot induce VDEs. Finally, $1/v_j$ describes the hypothetical depth values used to sample $I$ along the negative disparity region of the epipolar line.

We made a critical observation to define $v_j$: relative VDE \enquote{motion} cannot be larger than the corresponding rigid flow generated by the scene depth and camera translation. That is, the disparity in VDEs is inversely proportional to the scene depths. This is further visualized in Fig. \ref{fig:vde_follow}-(c) and (d) which display that the disparity relative to the reflective surface projection is larger for the closer distance $z_2$ than that at a distance $z_1$ in Fig. \ref{fig:vde_follow}-(b). Then, instead of modeling $|v_j|$ to vary from 0 to $1/t_n$, we define it as
\begin{equation} \label{eq:vde_n}
v_j = -\tfrac{j}{N_v-1}\left(\tfrac{1}{\hat{D}}-\epsilon\right)-\epsilon,
\end{equation}
where $\epsilon$ is a very small number. $V^P$ in Eq. (\ref{eq:vde_syn}) is the projected VDE probabilities from the VDE logits $V^L$ given by
\begin{equation} \label{eq:vde_vp}
V^{P}(\bm{p}) = \sigma \left(\{V^L_j(g(\bm{p}, v_j, \mathbf I|\bm{t}_c, K))\}^{N_v-1}_{j=0} \right).
\end{equation}

Similar to $\hat{D}$ in Eq. (\ref{eq:depth_hat}), a VDE activation map $\hat{V}$, which is useful for visualizing the most reflective regions in an image, can then be obtained by
\begin{equation} \label{eq:v_hat}
\hat{V}(\bm{p}) = \{v_j\}^{N_v-1}_{j=0} \cdot \sigma(V^L(\bm{p})).
\end{equation}

\subsection{NVSVDE-Net}

With our relaxed volumetric rendering (Eq. \ref{eq:vre_approx}) and VDE synthesis (Eq. \ref{eq:vde_syn}), we can generate novel views with VDEs via the respective depth (geometry) and VDE logits, $D^L$ and $V^L$. Fig. \ref{fig:overall_arch} depicts the overall architecture of our NVSVDE-Net for joint learning of NVS and VDEs. The NVSVDE-Net predicts $D^L$ and $V^L$ from geometry and VDE pixel-aligned features in \textit{a single forward pass} by
\begin{equation}\label{eq:DL_VL}
\begin{split}
D^L(\bm{p}) &= F_{D}(W_D(\bm{p}), \gamma(\bm{p}, R_c, \bm{t}_c)), \\
V^L(\bm{p}) & = F_{V}(W_V(\bm{p}), \gamma(\bm{p}, R_c, \bm{t}_c)), \\
\end{split}
\end{equation}
where $F_D$ and $F_V$ are MLP heads (Linear-ELU-Linear) that re-calibrate the geometry and VDE pixel-aligned features, $W_D$ and $W_V$ respectively, for the target camera view $c$. Note that such calibration is needed as the samples at distances $t_i$ (in Eq. \ref{eq:vre_approx}) and $1/v_i$ (in Eq. \ref{eq:vde_syn}) are measured from the target camera view $I_c$, not the input source view $I$. Pixel positional information $\bm{p}$ and relative camera extrinsics $R_c$ and $t_c$ are also fed into $F_D$ and $F_V$ via the positional encoding $\gamma(\cdot)$. We explored both a learnable $\gamma_{\theta}$ (fast and compact but can potentially overfit) and a sine-cosine \cite{nerf} $\gamma$ positional encoding (more general but slower due to requiring more channels) where there was little difference.

$W_D$ and $W_V$ enable cross-scene generalization and are simultaneously estimated from a CNN-based (or transformer-based) encoder-decoder backbone $F_W$ as shown in Fig. \ref{fig:overall_arch}. $F_W$ is also fed with the single image input $I$ and the relative pixel locations $(U,V)$ to improve learning from random-resized and -cropped patches \cite{pladenet} as $\left[W_D, W_V\right] = F_W(I, (U, V))$.

Once $D^L$ and $V_L$ are computed, they can be used in Eqs. (\ref{eq:vre_approx}) and (\ref{eq:vde_syn}) to yield the VDE-infused synthetic view $I''_c$, as shown in the center of Fig. \ref{fig:overall_arch}. 
However, the quality of $I''_c$ is closely tied to the number of samples $N$ in our relaxed volumetric rendering approximation. Naively increasing $N$ can incur additional computational complexity. Instead, we incorporate a sampler block $F_S$ to estimate fine-grained ray samples from projected depth probabilities and colors.

\enlargethispage{\baselineskip}
\subsubsection{The Sampler Block}

The sampler block $F_S$ in our NVSVDE-Net (top left of Fig. \ref{fig:overall_arch}) takes as input the projected probability logits $D^P$ and projected VDE-infused colors from $I^v_c$. These are then mapped by a fully-connected network (Linear-ELU-Linear-ELU-Linear) into $N^*$ refined per-pixel sampling distances $\bm{t}^*_k(\bm{p})$ and soft-maxed weights $\bm{w}^*_k(\bm{p})$ as given by
\begin{equation} \label{eq:ft3_vr}
\bm{t}^*_k(\bm{p}), \bm{w}^*_k(\bm{p}) = F_S(D^P(\bm{p}), \{I^V_c(g(\bm{p}, t_i, R_c|t_c, K))\}^{N-1}_{i=0}).
\end{equation}
$\bm{w}^*_k(\bm{p})$ and $\bm{t}^*_k(\bm{p})$ respectively replace $D^P_i$ and $t_i$ in Eq. (\ref{eq:vre_approx}) to yield the final synthetic image $I'_c$ by a \textit{fine-grained} relaxed volumetric rendering as
\begin{equation} \label{eq:final_synth}
I'_c(\bm{p}) = \textstyle\sum^{N^*}_{k=1} \bm{w^*_k(\bm{p})} I^v_c(g(\bm{p}, t^*_k(\bm{p}), R_c|\bm{t}_c, K).
\end{equation}

% \subsection{Differences With Previous methods}

\noindent Note that the network architecture in Fig. \ref{fig:overall_arch} only requires the backbone to be \textit{run once per-reference image}. Once $W_D$ and $W_V$ are estimated, the novel views are generated by running the computationally inexpensive re-calibration ($F_D$, $F_V$) and the sampler ($F_S$) blocks according to the relaxed VRE in Eq. (\ref{eq:final_synth}). Contrary to previous works \cite{pixelnerf, behindScenes, scenerf, mine} that also incorporate MLP heads for late-stage rendering, we only need to run them once instead of running them for each point in the target rays. Furthermore, previous single-view-based NVS methods \cite{mine, pixelnerf, behindScenes, scenerf, single_view_mpi} require either depths or pose GTs (or both), while our method learns from image sequences only in an entirely self-supervised manner with the aid from an improved camera pose estimation network.

\begin{figure}
  \centering 
  \includegraphics[width=0.48\textwidth]{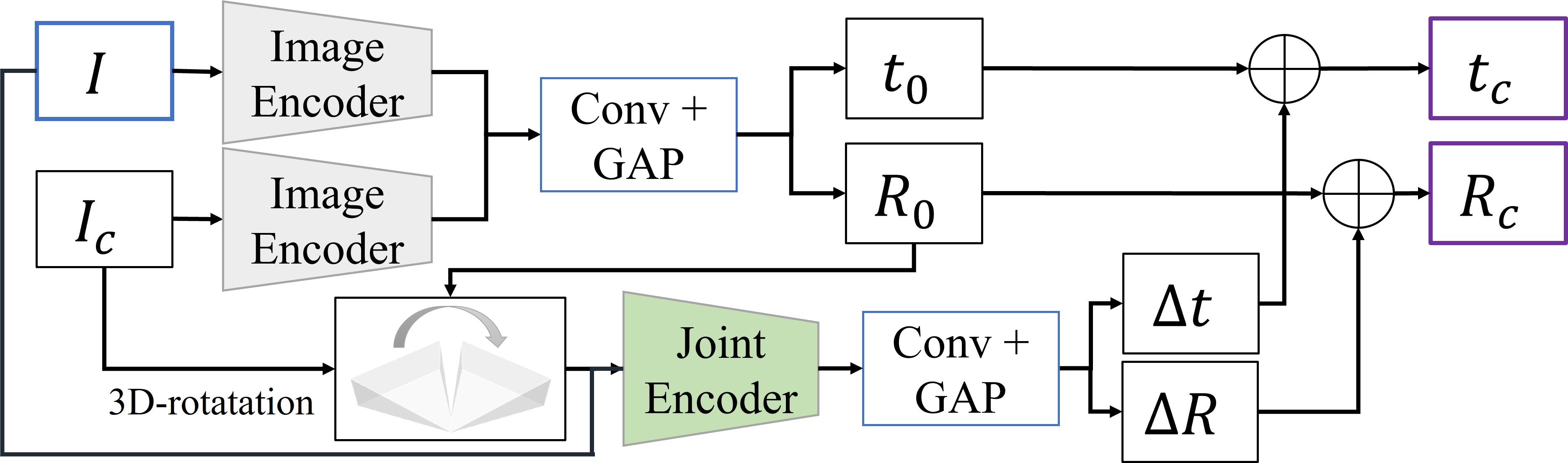}
  \vspace*{-5mm}
  \caption{Improved PoseNet. Our PoseNet refines initial coarse extrinsics by predicting a residual between rotation-aligned views.}
  % Poses are predicted between $I$ and $I_{-1}$ and $I_{+1}$, respectively.}
  \label{fig:posenet}
\end{figure}

\subsection{Improved Camera Pose Estimation}

Previously, camera-pose estimation networks \cite{sfmlearner, monodepth2} have shown reasonable performance for the monocular depth estimation task in driving datasets \cite{kitti2012, cityscapes}. However, they can only handle relatively simple camera motions because the cameras mounted on cars are primarily exposed to Z-axis motion and small horizontal rotations. In contrast, indoor datasets \cite{mpi} and hand-held captured scenes \cite{MCData} contain arbitrary translations with considerable rotations on all axes.

We observe that based on perceptual features, it is much simpler to understand 3D translation between two images if they are first rotation-aligned. We incorporate this observation into our improved PoseNet as depicted in Fig. \ref{fig:posenet}. In our PoseNet, from a pair of input images ($I$, $I_c$), coarse camera extrinsics ($R_0$, $\bm{t}_0$) are estimated by aggregating their deep features (estimated by a shared encoder network) with a convolutional and global average pooling (Conv+GAP) layer. This coarse stage resembles the PoseNets in \cite{sfmlearner, monodepth2}. $R_0$ is then used to rotation-align $I_c$ to $I$. We then extract deep features from the rotation-aligned pair that contain much more relevant visual features, such as closer vanishing points and disparities. By an additional Conv + GAP layer, residual translation and rotation values, $\Delta R$ and $\Delta \bm{t}$, are computed and added to $R_0$ and $\bm{t}_0$ to yield the final relative extrinsic parameters $R_c$ and $\bm{t}_c$.

\subsection{Loss Functions}
\label{sec:losses}

From a training input image $I$, synthetic images are estimated for both the previous and next views ($I_{-1}$ and $I_{+1}$) in the video sequence.
We utilize a synthesis loss, $l_{syn}$, between each synthetic view, both coarse and fine ($I''_{-1}$ and $I''_{+1}$, $I'_{-1}$ and $I'_{+1}$, respectively) and their corresponding GT images ($I_{-1}$ and $I_{+1}$). Additionally, we incorporate disparity smoothness $l_{sm}$ that aids in regularizing $\hat{D}$. The total loss function is given by
\begin{equation} \label{eq:total_loss}
l_{total} = l_{syn}(I''_c) + l_{syn}(I'_c) + \alpha_{sm}l_{sm}.
\end{equation}

\vspace*{1mm}
\noindent
\textbf{The synthesis loss,} $l_{syn}$, is a combination of L1 and perceptual (VGG) loss \cite{perceptual} to enforce similar colors and structures between synthetic and GT views. The VGG loss is useful for penalizing deformations, textures, and lack of sharpness as it compares estimated and GT views in the deep feature space of a pre-trained image classification network. The L2 norm of the perceptual error of the first three max-pooling layers from the $VGG19$ \cite{vgg}, denoted by $\phi^l$, was utilized. $l_{syn}$ is given by
\begin{equation} \label{eq:l_syn}
l_{syn} = ||I^o_c - I_c||_1 + \alpha_p\textstyle\sum_{l=1}^{3} ||\phi^l(I^o_c)-\phi^l(I_c)||^2_2,
\end{equation}
where $\alpha_p=0.01$ balances the contributions of the L1 and VGG terms. $I_c^o=(1 - O_c )\odot I_c  + O_c\odot I'_c$ is the synthetic view with occluded contents replaced by those in $I_c$. $\odot$ is the Hadamart product. We compute the occlusion mask $O_c$ following \cite{llff, falnet} as
\begin{equation} \label{eq:synth_free}
O_c(\bm{p}) = \textstyle\sum_{i=0}^{N-1} \sigma(D^L)_i(g\left(\bm{p}, t_i,  R_c|t_c, K\right)),
\end{equation}

\textbf{The edge-aware smoothness} loss $l_{sm}$ with a weight of $a_{sm}=0.05$ regularizes $\hat{D}$ to be smooth in homogenous image regions. See \textit{Supplemental} for more details.
\enlargethispage{\baselineskip}
\section{Experiments and Results}
\label{sec:experiments_results}

Extensive experimental results show our method can generate, for the first time, NVS with VDEs from single image inputs on real datasets that contain complex camera motions and scenes with no depth or pose annotations. Please see \textit{Supplemental} for additional results and \textit{videos}.

% \subsection{Implementation Details}
We train our NVSVDE-Net with the Adam optimizer ($\beta_1$: 0.9, $\beta_2$: 0.999) with a batch size of 6 for 50 epochs and 3k iterations per epoch. 
The initial learning rate is set to $10^{-4}$ and is halved at 50, 75, and 90\% of the training for stability and convergence.
We set $N=32$, $N_v=32$, and $N^*=16$ for our relaxed volumetric rendering. 
We train all models on the RealEstate10k \cite{mpi} and MannequinChallenge \cite{MCData} datasets with random spatial data augmentations such as random resize and crops, random horizontal flip, and value-based data augmentations such as random gamma, brightness, and color shifts. Training patches are of 240$\times$426 obtained from randomly resized images between 30\% and 85\% of their original resolution. Finally, 8-bit RGB images are normalized to the $[-0.5, 0.5]$ range. 

We adopt the same ResNet34 \cite{resnet} IMAGENet \cite{imagenet} pre-trained backbone in all models (in ours and \cite{pixelnerf, behindScenes, scenerf}), otherwise specified. In particular, we set $N$ = 64, 48, and  32 ray samples for PixelNeRF \cite{pixelnerf}, BehindScenes \cite{behindScenes}, and SceneRF \cite{scenerf}, respectively. To match NVSVDE-Net's $N^*$, we use 4 Gaussians with four samples in SceneRF. For a fair comparison, we train \cite{pixelnerf, behindScenes, scenerf} under the same self-supervised conditions as our NVSVDE-Net. We observed \cite{pixelnerf, behindScenes, scenerf} are unstable in the early epochs when learning self-supervised, so they are trained from a fully-trained PoseNet from our best NVSVDE-Net.

\enlargethispage{\baselineskip}
\subsection{Datasets}

\textbf{RealEstate10k (RE10) \cite{mpi}} is a dataset consisting of 10M frames from approximately 80k video clips from around 10k YouTube videos of indoor and outdoor human-made environments. 
After downloading, we account for 400k and 180k for training and testing, respectively. We randomly select the 4, 8, 12, or 16$^{th}$ previous or next frame for training. We use every 1k sample from the test set for testing.

The \textbf{MannequinChallenge (MC)} \cite{MCData} dataset is a large, diverse, and challenging dataset that mainly contains humans in both indoor and outdoor scenarios. It contains ~170K frames from 2k YouTube videos where people try to stay stationary while a recording camera moves around the scene. After downloading, we account for 60k samples for training and 9k samples for testing. We randomly select up to the 6$^{th}$ previous or next frame for training. We use every 20$^{th}$ frame from the test set for testing.

In both datasets, images have a full resolution of 720×1280. We train with 1/3 resolution patches and test with 1/2 resolution inputs. We render two views from an input test view, the previous $k$-th frame and the next $k$-th frame for evaluation. $k$ is set to 8 and 1 for RE10k and MC, respectively. Network outputs are bilinearly up-scaled to full resolution before measuring their quality metrics. We measure the quality of the rendered novel views with RMSE, PSNR, SSIM \cite{ssim}, and LPIPS \cite{lpips} metrics.

\begin{table}
    \scriptsize
    \centering
    \setlength{\tabcolsep}{3pt}
    \renewcommand{\arraystretch}{1.2}
\begin{tabular}{lcccccc}
\hline
Methods & VDE & MAE$\downarrow$ & PSNR$\uparrow$ & PSNR$_{lf}$$\uparrow$ & SSIM$\uparrow$ & LPIPS$\downarrow$ \\ 
\hline
\multicolumn{7}{c}{RealEstate10k (RE10k) \cite{mpi} Dataset \cite{mpi}} \\

PixelNerf \cite{pixelnerf} & No & 0.0417 & 22.8455  & 28.0945 & 0.7818 & 0.3256 \\

BehindScenes \cite{behindScenes} & No & 0.0466 & 22.9949  & 28.5941 & 0.8068 & 0.2762 \\

SceneRF \cite{scenerf} & No & 0.0373 & 23.6087  & 28.9636 & 0.8130 & 0.2709 \\

\textbf{NVSVDE-Net (Ours)} & \textbf{Yes} & \textbf{0.0319} & \textbf{24.3131}  & \textbf{30.2529} & \textbf{0.8397} & \textbf{0.2325} \\

\hline
\multicolumn{7}{c}{MannequinChallenge (MC) Dataset \cite{MCData}} \\

PixelNerf \cite{pixelnerf} & No & 0.0511 & 21.3047  & 25.2781 & 0.7580 & 0.3455 \\

BHindScenes \cite{behindScenes} & No & 0.0463 & 21.4307  & 25.9280 & 0.7831 & 0.3101 \\

SceneRF \cite{scenerf} & No & 0.0467 & 21.5992 & 25.8119 & 0.7796 & 0.3080 \\

\textbf{NVSVDE-Net (Ours)} & \textbf{Yes} & \textbf{0.0405} & \textbf{22.4274}  & \textbf{27.0263} & \textbf{0.8130} & \textbf{0.2733} \\

\hline
\end{tabular}
    \vspace{-3mm}
    \caption{NVS results. $\downarrow$/$\uparrow$ denotes the lower/higher, the better.}
    \label{tab:results}
\end{table}

\begin{table}
    \scriptsize
    \centering
    \setlength{\tabcolsep}{2pt}
    \renewcommand{\arraystretch}{1.2}
\begin{tabular}{lcccccc}
\hline
Methods & VDE & MAE$\downarrow$ & PSNR$\uparrow$ & PSNR$_{lf}$$\uparrow$ & SSIM$\uparrow$ & LPIPS$\downarrow$ \\ 
\hline
\multicolumn{7}{c}{RealEstate10k (RE10k) Dataset \cite{mpi}} \\
% VDE
(w/o VDE) & No & 0.0323 & 24.1679  & 30.1651 & 0.8362 & 0.2320 \\
(VDE disabled) & No & 0.0323 & 24.2092  & 30.1046 & 0.8383 & 0.2325 \\

% Relaxed vol rend
(w/o $F_D$ \& $F_V$) & No & 0.0337 & 23.9710  & 29.8303 & 0.8291 & 0.2426 \\
($N$= 48, $N^*$= 0) & Yes & 0.0332 & 23.9613  & 29.8602 & 0.8322 & 0.2417 \\
($N=32$, $N^*$= 32)& Yes & 0.0321 & 24.3007  & 30.2156 & 0.8389 & 0.2340 \\
($I''_c$) & Yes & 0.0325 & 24.1020  & 29.9808 & 0.8343 & 0.2365 \\
($\sigma$-based VRE) & Yes & 0.0325 & 24.1729  & 30.1202 & 0.8361 & 0.2375 \\
(Periodic $\gamma$ \cite{nerf}) & Yes & 0.0323 & 24.2506  & 30.1848 & 0.8366 & 0.2350 \\

(No resize-crop, test at 1/3) & Yes & 0.0331 & 24.1030 & 30.4413 & 0.8213 & 0.2672 \\

% Backbones
(R18)  & Yes & 0.0338 & 24.0340  & 29.9165 & 0.8289 & 0.2437 \\
(Swin-t \cite{swint})  & Yes & 0.0322 & 24.1917  & 30.1424 & 0.8354 & 0.2365 \\

(\cite{sfmlearner}'s PoseNet) & Yes & 0.0337 & 23.8736  & 29.8197 & 0.8259 & 0.2415 \\
(Full) & Yes & 0.0319 & 24.3131  & 30.2529 & 0.8397 & 0.2325 \\

\hline
\multicolumn{7}{c}{MannequinChallenge (MC) Dataset \cite{MCData}} \\

(\cite{sfmlearner}'s PoseNet) & Yes & 0.0454 & 21.5004  & 26.2065 & 0.7787 & 0.2897 \\

(Full) & Yes & 0.0405 & 22.4274  & 27.0263 & 0.8130 & 0.2733 \\

\hline
\end{tabular}
    \vspace{-3mm}
    \caption{NVSVDE-Net ablation studies.}
    \label{tab:ablations}
\end{table}

\begin{figure}[t]
  \centering 
  \includegraphics[width=0.48\textwidth]{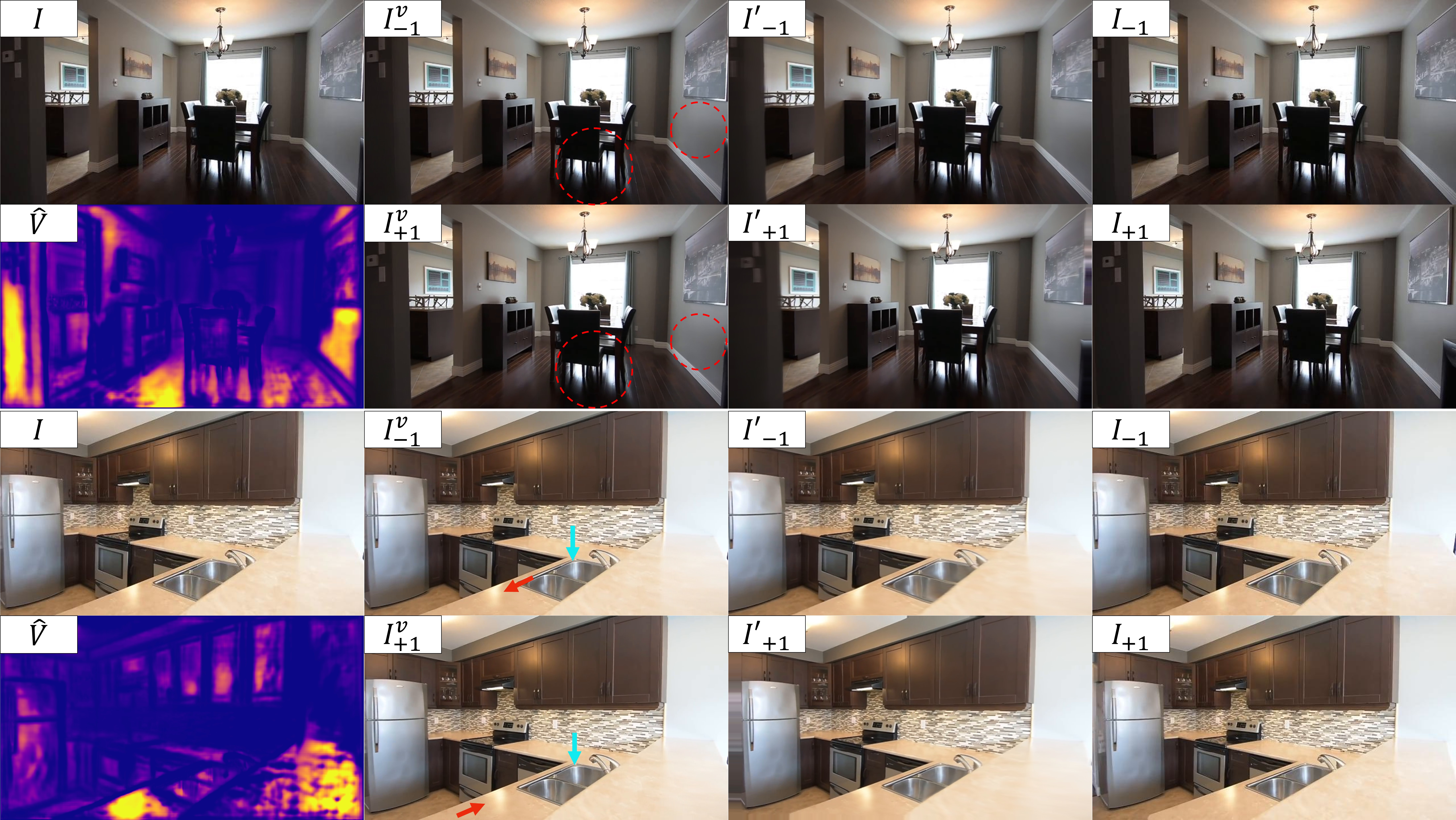}
  \vspace*{-6mm}
  \caption{Our NVSVDEnet yields novel views with VDEs. See \url{https://shorturl.at/ltJT7} Fig5-video.}
  \label{fig:re10k_vde}
  \vspace{-3mm}
\end{figure}

\begin{figure}[t]
  \centering 
  \includegraphics[width=0.49\textwidth]{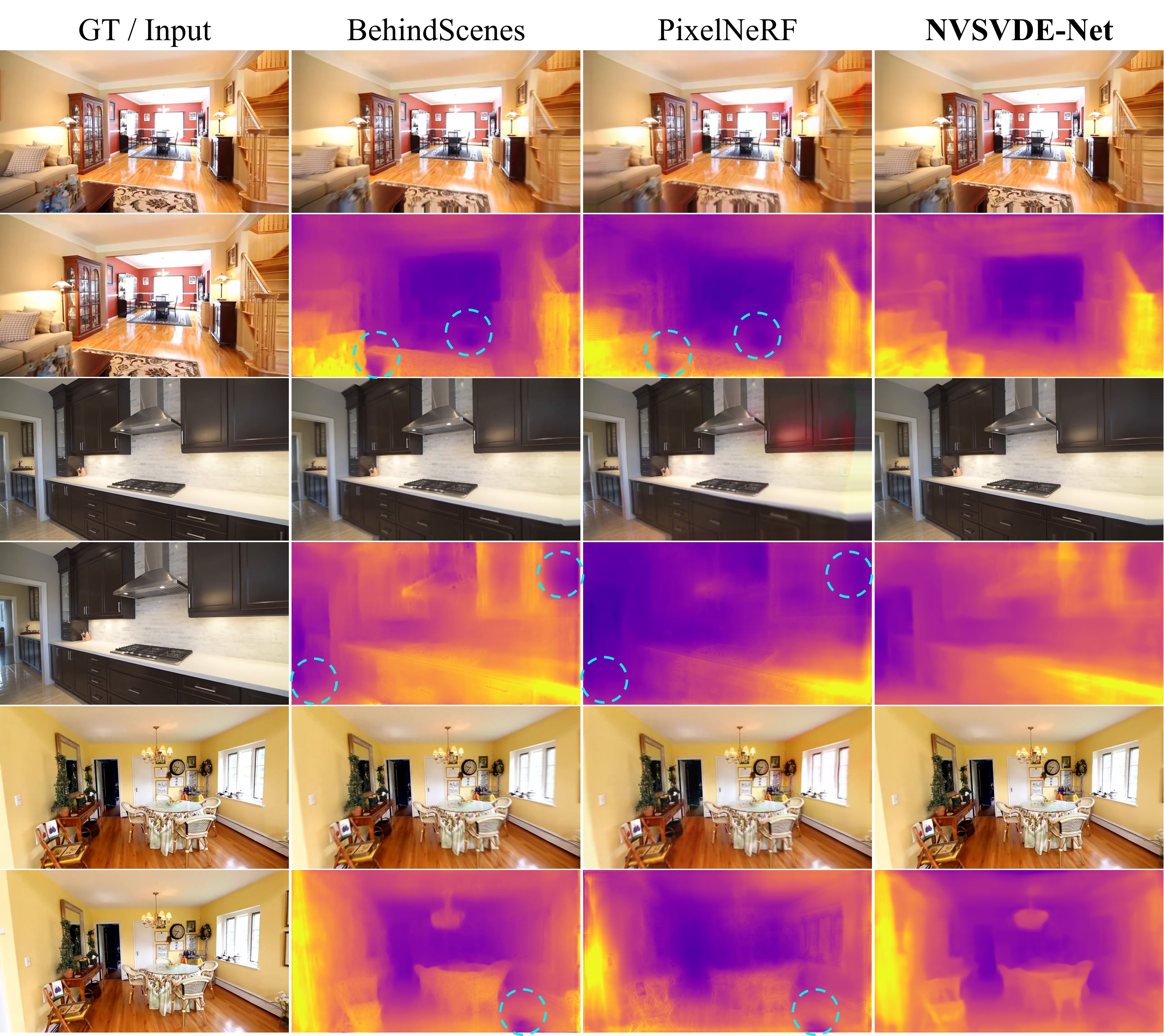}
  \vspace*{-6mm}
  \caption{Comparison with previous methods on RE10 \cite{mpi}. See \url{https://shorturl.at/ltJT7} Fig6-video}
  \label{fig:re10_nvs}
\end{figure}

\enlargethispage{\baselineskip}
\subsection{Results on RealEstate10k (RE10k)}

The scenes in RE10k contain a considerable amount of perceptually significant view-dependent effects. As shown in Fig. \ref{fig:re10k_vde}, our NVSVDE-Net learns single-view-based realistic NVS and VDEs whose activations map the most reflective regions in the corresponding input images well. 
% Even when VDEs are realistic, there is still room for improvement. 

\textbf{Measuring VDEs} is difficult due to their sparse nature. We observed that most VDE that could be modeled from single images consist of low-frequency information, such as glossy reflections. Based on that assumption, we propose the PSNR$_{lf}$ metric in Table \ref{tab:results}. PSNR$_{lf}$ takes the PSNR between an estimated image and its GT image, both filtered by a $21 \times 21$-sized Gaussian kernel before comparison. Other metrics that focus on strong image structures, such as SSIM and LPIPS, showed transparency to the changes in VDEs, as VDEs rarely change the high-frequency information or high-level structures in the natural scenes. 
\enlargethispage{\baselineskip}

Table \ref{tab:results} shows the results on RE10K \cite{mpi}. Our NVSVDE-Net \textit{significantly} outperforms SceneRF \cite{scenerf} by a large margin of \textbf{0.7dB} in PSNR and \textbf{1.3dB} in PSNR$_{lf}$ and previous methods \cite{pixelnerf, behindScenes} by more than \textbf{1.3dB} in PSNR and \textbf{1.6dB} in PSNR$_{lf}$. We qualitatively compare PixelNeRF, BehindScenes, and NVSVDE-Net in Fig. \ref{fig:re10_nvs}, which shows that our model yields more detailed synthetic views with fewer artifacts. Fig. \ref{fig:re10_nvs} also shows that methods \cite{pixelnerf, behindScenes} that do not model VDEs struggle to yield plausible geometries for the reflective regions.

% Ablate: 
% Mainly (as these are the contributions)
% 1. VDE
%    * w/o VDE
%    * VDE disabled
% 2. Relaxed volumetric rendering
%    * Using sigma estimation instead of W
%    * Only coarse with more N
%    * Coarse output I''_c
% 3. Self-supervised stuff
%    * w/o improved posenet
%    * w/o supp on I''c
% Secondarily
% 4. Network design choices
%    * use of F_D and F_V
%    * different backbones 

\enlargethispage{\baselineskip}
\subsubsection{Ablation Studies}
\label{sec:fvs_ablation}
\begin{figure}[t]
  \centering 
  \includegraphics[width=0.49\textwidth]{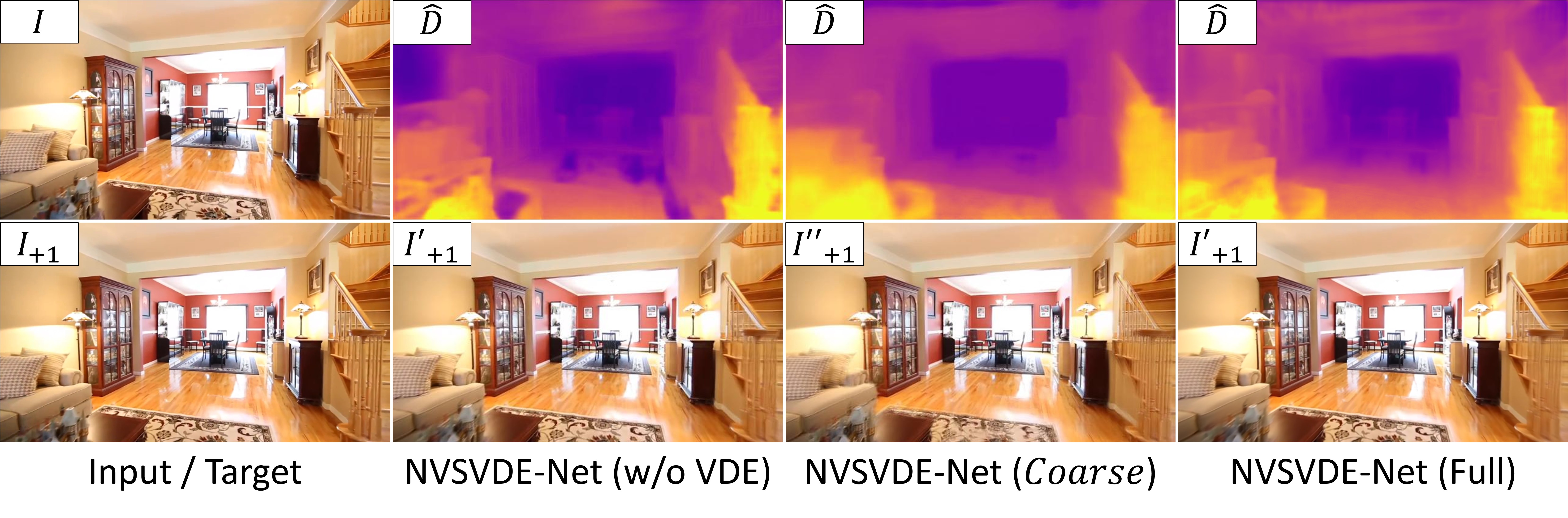}
  \vspace*{-8mm}
  \caption{NVSVDE-Net ablation studies. See \url{https://shorturl.at/ltJT7} Fig8-video.}
  \label{fig:ablations}
\end{figure}

Table \ref{tab:ablations} shows ablation studies for our NVSVDEnet. We study the effects of our proposed VDE synthesis in (w/o VDE), which yields 0.15dB lower PSNR than our full model. Even though the PSNR difference is not very large, a clear advantage in predicted geometries can be observed in Fig. \ref{fig:ablations}, where our model w/o VDEs predicts holes for the reflective regions. We also ablate the effects of our VDEs by disabling the VDE estimation in our full model, denoted by (VDE disabled), which yields 0.15dB lower PSNR$_{lf}$. 

Next, we ablate design choices in our proposed relaxed volumetric rendering. (w/o $F_D$ \& $F_V$) in Table \ref{tab:ablations} shows the negative effects of not adjusting the depth and VDE logits volumes. ($N=48$, $N^*=0$) shows that the coarse synthetic output, even with more samples, is still 0.3dB in PSNR behind our full model. On the other hand, ($N=32$, $N^*=32$) shows that only a few adaptive $N^*=16$ in our full model is enough for optimal results. ($I''_c$) is the coarse synthetic output of our full model. Interestingly, ($I''_c$) is 0.2dB lower than $I'_c$, but 0.14bB higher in PSNR than ($N=48$, $N^*=0$), showing that the joint supervision of $I'_c$ benefits it. The fine-grained $I'_c$ removes most double-edge artifacts due to discrete ray distance discretization as shown in Fig. \ref{fig:ablations}. In our relaxed volumetric rendering, we also explore estimating density values $\sigma$  instead of $\mathcal{T} (t)\sigma(t)$. However, estimating $\sigma$ was not only computationally more expensive due to the computation of the accumulated transmittance for each ray sample but also yielded slightly worse quality metrics. A periodic encoding $\gamma$ was also explored, but we observed slightly worse quality metrics.

We also provide metrics for our NVSVDE-Net with different backbones, such as ResNet18 (R18) and Swin tansformer \cite{swint} (Swin-t) transformer backbone. Even when NVSVDE-Net (swint) quantitative metrics are not better our ResNet34 full model, its estimated depth and VDE maps are perceptually more consistent (see \textit{Supplemental} for qualitative comparison).

Finally, we also ablate the effect of our improved PoseNet, with a synthesis quality impact of $\sim$0.5dB and $\sim$0.9dB higher PSNR than the widely adopted \cite{sfmlearner}'s PoseNet, on RE10k and MC respectively. Test poses are also estimated, reflecting the quality of our PoseNet.

% which also provides a global scale, removing the need for pose and point-cloud annotations for scale-invariant training \cite{single_view_mpi}.

\begin{figure}[t]
  \centering 
  \includegraphics[width=0.49\textwidth]{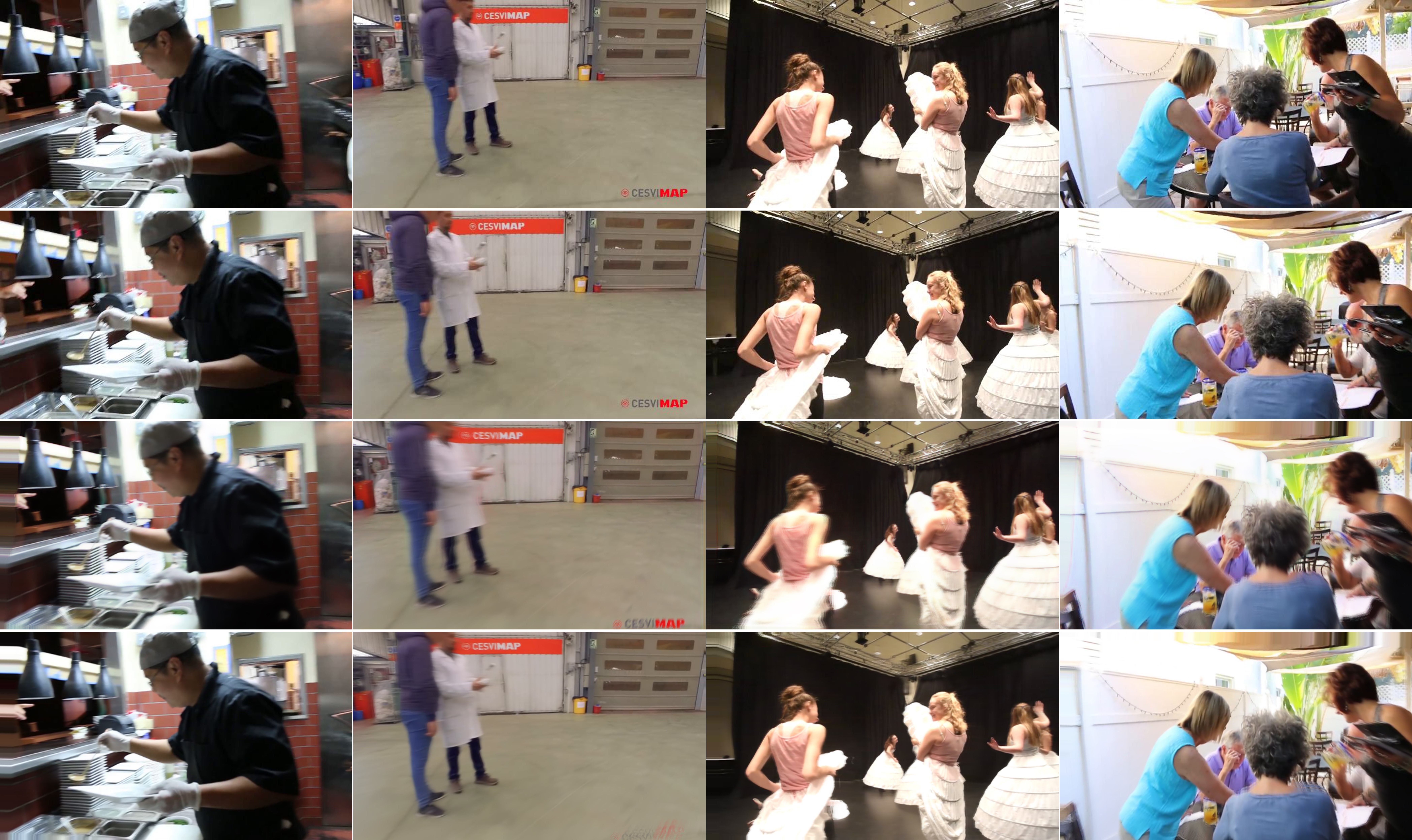}
  \vspace*{-6mm}
  \caption{Results on MC \cite{MCData}. Top to bottom: $I$, GT, SecneRF \cite{scenerf}, Ours. See \url{https://shorturl.at/ltJT7} Fig7-video.}
  \label{fig:mc_nvs}
\end{figure}

\enlargethispage{\baselineskip}
\subsection{Results on MannequinChallenge (MC)}

The bottom part of Table \ref{tab:results} shows quantitative results for single-image-based NVS on the MC \cite{MCData} validation set. The NVSVDE-Net outperforms the existing methods at least by \textbf{0.8dB} in PSNR and \textbf{$\sim$1.2dB} in PSNR$_{lf}$. Fig. \ref{fig:mc_nvs} depicts qualitative comparisons between our NVSVDE-Net and SceneRF \cite{scenerf}. Again, our method yields sharper and more consistent single-view NVS outputs.

\subsection{Runtime Analysis}

Our NVSVDE-Net takes 42ms for a 640$\times$360 image on an Nvidia A6000 GPU on vanilla PyTorch. 12ms are used for the backbone operation and 30ms for projections, sampler, and rendering. Since the backbone runs only once for each input image, it only takes 30ms to render novel views of the same scene. Even though our method is near real-time with 33FPS, further speed-ups are possible with optimized Python packages, such as TensorRT, which can provide up to 3$\times$ faster rendering. In comparison, the other methods in Table \ref{tab:results}, which all share the same 25M-parameter backbone, take 160, 112, and 99ms for rendering a 640$\times$360 image for SceneRF, PixelNeRF, and BehindScenes, respectively.
\enlargethispage{\baselineskip}
\section{Conclusions}
\label{sec:conclusions}

We firstly presented a novel method, the NVSVDE-Net, that can learn to perform NVS and VDE estimation on single images in a self-supervised manner from monocular image sequences. We showed that our method generalizes well on unseen test images and that it can generate plausible VDEs and depth maps from a single image. 
In addition, our NVSVDE-Net incorporates a relaxed approximation to volumetric rendering, which we further improve by incorporating a sampler module for fine-grained ray sampling and rendering. Our NVSVDE-Net yields more realistic NVS images with VDEs in comparison to the recent SOTA methods such as PixelNeRF, BHindScenes, and SceneRF on the RE10k and MC datasets. 
\enlargethispage{\baselineskip}

\section*{Acknowledgements}
This work was supported by IITP grant funded by the Korea government (MSIT) (No. RS2022-00144444, Deep Learning Based Visual Representational Learning and Rendering of Static and Dynamic Scenes).

{
    \small
    \bibliographystyle{ieeenat_fullname}
    \bibliography{main_submission}
}


\end{document}